%% file: main.tex
\documentclass[sigconf, twocolumn]{acmart}
\AtBeginDocument{%
  }

\copyrightyear{2025}
\acmYear{2025}
\setcopyright{acmlicensed}
\acmConference[MM '25] {Proceedings of the 33rd ACM International Conference on Multimedia}{October 27--31, 2025}{Dublin, Ireland.}
\acmBooktitle{Proceedings of the 33rd ACM International Conference on Multimedia (MM '25), October 27--31, 2025, Dublin, Ireland}
\acmISBN{979-8-4007-2035-2/2025/10}
\acmDOI{10.1145/3746027.3754910}

\usepackage{makecell}
\usepackage{bm}

\begin{document}

\title{InteractMove: Text-Controlled Human-Object Interaction Generation in 3D Scenes with Movable Objects}


\author{Xinhao Cai}
\affiliation{%
  \institution{Wangxuan Institute of Computer\\  Technology, Peking University}
  \city{Beijing}
  \country{China}}
\email{xinhao.cai@stu.pku.edu.cn}

\author{Minghang Zheng}
\affiliation{%
  \institution{Wangxuan Institute of Computer\\ Technology, Peking University}
  \city{Beijing}
  \country{China}}
\email{minghang@pku.edu.cn}

\author{Xin Jin}
\affiliation{%
  \institution{Beijing Electronic Science and Technology Institute}
  \city{Beijing }
  \country{ China}}
\email{jinxinbesti@foxmail.com }

\author{Yang Liu}
\authornote{Corresponding author}
\affiliation{%
  \institution{Wangxuan Institute of Computer Technology}
  \institution{State Key Laboratory of General Artificial Intelligence, Peking University}
  \city{Beijing}
  \country{China}}
\email{yangliu@pku.edu.cn}

\renewcommand{\shortauthors}{Xinhao Cai, Minghang Zheng, Xin Jin, and Yang Liu.}
\newcommand{\MethodName}{Affordance-Guided Collision-Aware Interaction Generation}
\newcommand{\MethodNameShort}{AGCA}

\begin{abstract}
In this paper, we propose a novel task of text-controlled human-object interaction generation in 3D scenes with movable objects. Existing human-scene interaction datasets suffer from insufficient interaction categories and typically only consider interactions with static objects (do not change object positions), and the collection of such datasets with movable objects is difficult and costly. To address this problem, we construct the InteractMove dataset for Movable Human-Object Interaction in 3D Scenes by aligning existing human-object interaction data with scene contexts, featuring three key characteristics: 1) scenes containing multiple movable objects with text-controlled interaction specifications (including same-category distractors requiring spatial and 3D scene context understanding), 2) diverse object types and sizes with varied interaction patterns (one-hand, two-hand, etc.), and 3) physically plausible object manipulation trajectories. With the introduction of various movable objects, this task becomes more challenging, as the model needs to identify objects to be interacted with accurately, learn to interact with objects of different sizes and categories, and avoid collisions between movable objects and the scene. To tackle such challenges, we propose a novel pipeline solution. We first use 3D visual grounding models to identify the interaction object. Then, we propose a hand-object joint affordance learning to predict contact regions for different hand joints and object parts, enabling accurate grasping and manipulation of diverse objects. Finally, we optimize interactions with local-scene modeling and collision avoidance constraints, ensuring physically plausible motions and avoiding collisions between objects and the scene. Comprehensive experiments demonstrate our method's superiority in generating physically plausible, text-compliant interactions compared to existing approaches. The code is available at \url{https://github.com/Cxhcmhhh/InteractMove}.

\end{abstract}




\maketitle

\begin{figure}
  \centering
  \includegraphics[width=\linewidth]{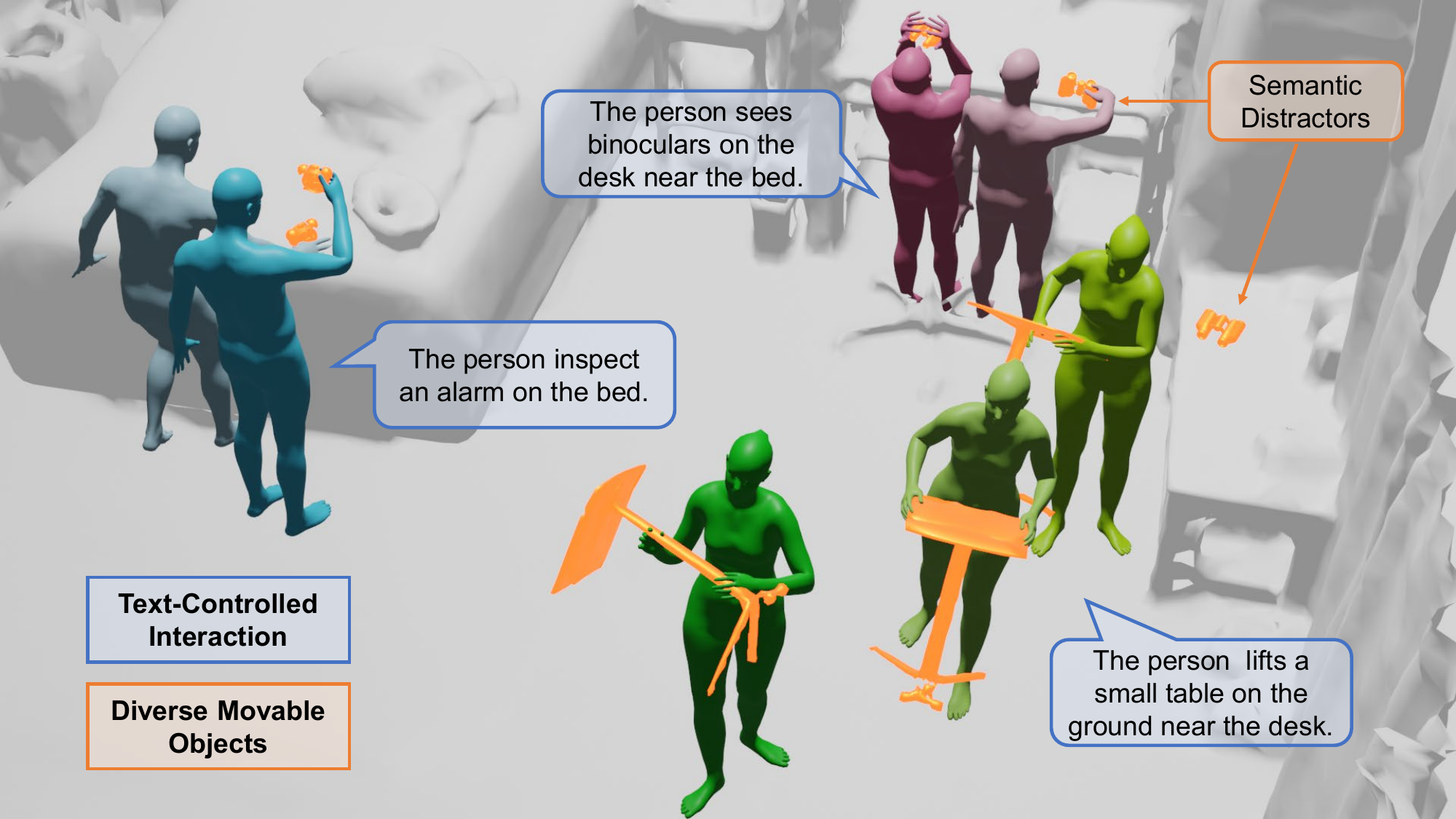}
  \caption{\label{fig_teaser} \textbf{Samples in our dataset.} We synthesize a large-scale human-object-interaction-in-scene dataset by aligning captured human-object interaction sequences with various 3D scan scenes. In the dataset, we provide free-form text annotations and interaction with movable objects in high-quality scenes.}
  \label{fig:teaser}
\end{figure}

\input{sec/1_intro}
\input{sec/2_rworks}
\input{sec/3_dataset}
\input{sec/4_method}
\input{sec/5_exp}
\input{sec/6_conclusion}

\bibliographystyle{ACM-Reference-Format}
\balance
\bibliography{sample-base}

\end{document}

%% file: sec/1_intro.tex
\section{Introduction}
\label{sec:intro}

The generation of human motions within scenes is a growing research area with significant applications in VR, AR, video games, and beyond. Recently, there has been increasing interest in generating human motions conditioned on natural language descriptions. However, most prior works either focus on language-driven interactions between humans and isolated objects~\cite{cghoi,hoidiff,t2hoi}, neglecting the influence of the surrounding scene, or study human-scene interactions~\cite{humanise,trumans} without explicitly considering movable objects. This results in limited expressiveness and practicality when deployed in real-world scenarios, where objects are often embedded in complex environments and exhibit various affordances. To bridge this gap, we propose a novel task: text-controlled human-object interaction generation in 3D scenes with movable objects.

In existing Human-Scene Interaction datasets~\cite{humanise,trumans,gtaim,pigraph,proxq}, interactions are quite limited, and interactable objects are often fixed and immovable, such as beds and sofas. Furthermore, manually collecting a new large-scale, high-quality 3D dataset is both difficult and costly. Therefore, we introduce the InteractMove dataset constructed by aligning existing human-object interaction data with richly annotated 3D scene contexts as shown in Fig.~\ref{fig:teaser}. Our dataset exhibits three key properties: (1) scenes contain multiple movable objects, enabling text-controlled interaction with specified targets, often in the presence of same-category distractors that necessitate spatial understanding; (2) the dataset covers diverse object types and sizes, with interactions that vary in complexity, including one-handed and two-handed actions; and (3) object manipulation trajectories are physically plausible, avoiding collisions. 

This new task also introduces the following challenges to be addressed. First, it requires models to comprehend natural language instructions and identify the correct object among multiple, often similar, distractors in the scene. For example, in the scene shown in Fig.~\ref{fig:teaser}, there are two binoculars, and the model needs to identify the one described in the text that is on the desk near the bed.
Second, the target objects may vary significantly in type and scale, exhibiting various affordances and requiring different interaction strategies. For example, lifting small objects like a cup may only require one hand to interact with, while larger objects like a table need both hands. Even for the same type of object, interaction strategies may differ depending on its specific shape, e.g., a cup with a handle is usually grasped by the handle, whereas a handleless cup is more naturally grasped by its body.
Third, the task involves dynamically manipulating objects within 3D scenes while ensuring physical plausibility, which includes avoiding penetrations or collisions with other scene elements, especially moving large objects over long distances with other crowded objects nearby. 

To address the above challenges, we propose a novel \MethodName{} (\MethodNameShort{}) framework with carefully designed components that model 3D object grounding, fine-grained hand-object joint affordance learning, and collision-aware motion generation. Specifically, we first employ state-of-the-art 3D visual grounding models to locate the intended object specified by the input text. To capture the diversity of object affordances and interaction strategies, we introduced a hand-object joint affordance learning module, which takes the object mesh as input and predicts the likelihood of interactions occurring between hand joints and object surfaces over time, referred to as hand-object affordance. This fine-grained affordance is used to guide the interaction motion generation, enabling more accurate interactions aligned with object size and interaction semantics. Finally, we incorporate a collision-aware motion generation strategy that voxelizes the region around the interactive object to evaluate spatial accessibility, combined with a collision-aware loss that enforces physically plausible motion and prevents interpenetration, while ensuring the object's trajectory remains synchronized with human control and scene constraints.

In summary, our contributions are threefold: (1) We introduce a new task that focuses on text-conditioned human-object interaction generation in movable-object 3D scenes. (2) We construct a comprehensive dataset for this task with text-controlled interaction and diverse movable objects; we also propose a novel framework for this task with carefully designed components that model 3D object grounding, fine-grained hand-object joint affordance learning, and collision-aware motion generation. (3) Extensive experiments demonstrate the effectiveness of our approach in producing realistic, text-aligned, and physically plausible interaction motions in 3D scene with movable objects.

%% file: sec/2_rworks.tex
\section{Related Works}
\label{sec:rwork}

\subsection{Human-Object Interaction }

Datasets capturing human-object interaction(HOI) are crucial for training generative models, yet remain difficult and costly to collect. Datasets like GRAB~\cite{grab}, BEHAVE~\cite{behave}, and CHAIRS~\cite{chairs} employed optical MoCap/IMU systems to capture detailed human-object interactions, including object trajectories. However, these interactions are typically performed in isolation without the presence of a full scene, thus lacking contextual constraints from the environment.  

Early works in this region begin with HOI detection\cite{hoidetect2023,hoidetect2024} or HOI image generation\cite{hoiimagegen}. For human-object interaction generation, early methods like OMOMO~\cite{omomo} rely on object trajectories as input, limiting their applicability in free-form generation tasks. InterDiff~\cite{interdiff} introduces object dynamics but focuses on motion prediction conditioned on past human motions. More recent works~\cite{hoidiff,cghoi,goal} attempt to generate interactions with isolated movable objects without the presence of a full scene.
Several recent methods also incorporate affordance prediction or contact map to inform interaction generation~\cite{affordance,t2hoi}. However, they generally model affordance as a coarse spatial heatmap over object surfaces, neglecting the affordance of hands. They neglect to explicitly model how different hand joints engage with object surfaces, which is important when differentiating between single-handed and two-handed interactions or fine-grained grasp strategies. In contrast, we propose hand-object joint affordance learning, which models fine-grained contact likelihoods between hand joints and object surfaces over time. This enables more accurate and diverse interaction generation aligned with object shape, size, and semantics.

\subsection{Human-Scene Interaction}

Earlier works begin with scene understanding\cite{sceneunderstanding2024}. Some recent efforts explore large-scale human-scene interaction (HSI). For instance, HUMANISE~\cite{humanise} synthesizes high-quality HSI data within virtual environments. Nonetheless, it supports only a limited range of immovable objects and predefined interaction types. 
TRUMANS\cite{trumans} is the dataset that is closest to us, as it also includes both scenes and dynamic objects. However, TRUMANS is limited in several key aspects: (1) the set of movable object categories is narrow (20 types), restricting interaction diversity; (2) actions are predefined into 10 coarse categories, lacking diverse interaction types; and (3) it only provide discrete action labels, lacking natural language annotations, thus unable to support language-guided HOI generation.
In contrast, our proposed dataset, InteractMove, extends beyond existing efforts by synthesizing realistic HOI data into richly annotated 3D scenes from ScanNet~\cite{scannet}, while enabling object movement and fine-grained textual control.  
It features 71 categories of movable objects embedded in 3D scenes, with interactions spanning 21 types and paired with natural language descriptions. This allows scene-aware and language-controllable interaction generation with dynamic objects, enabling more realistic and physically plausible interactions in complex environments.

Human-scene interaction generation uses conditional Variational Auto-Encoder (cVAE)\cite{cvae} or diffusion models\cite{diffusion} to generate human-scene interaction based on action labels or text conditions. Earlier works~\cite{singleframe1,singleframe2,singleframe3} mainly focus on predicting static human poses conditioned on the scene geometry, often for single frames. Later approaches~\cite{mulframe1,mulframe2,mulframe3} extended this to temporal sequences, enabling more realistic interactions. However, these methods typically rely on action labels or scene cues, without leveraging natural language instructions.
HUMANISE~\cite{humanise} proposes text-guided motion generation within static scenes. Yet, it does not handle dynamic object manipulation, as all interactable items are fixed. This severely limits interaction diversity and realism. With the introduction of various movable objects, this task becomes more challenging, as the model needs to accurately identify interactive objects, learn to interact with objects of different sizes and categories, and avoid collisions between objects and the scene. Our work addresses these challenges by introducing a text-controlled generation framework that incorporates 3D object grounding, fine-grained hand-object joint affordance learning, and collision-aware motion generation, enabling precise, diverse, and realistic interactions in complex scenes.

%% file: sec/3_dataset.tex
\section{InteractMove Dataset}
\label{dataset}

\begin{figure}
\centering
\includegraphics[width=0.8\linewidth]{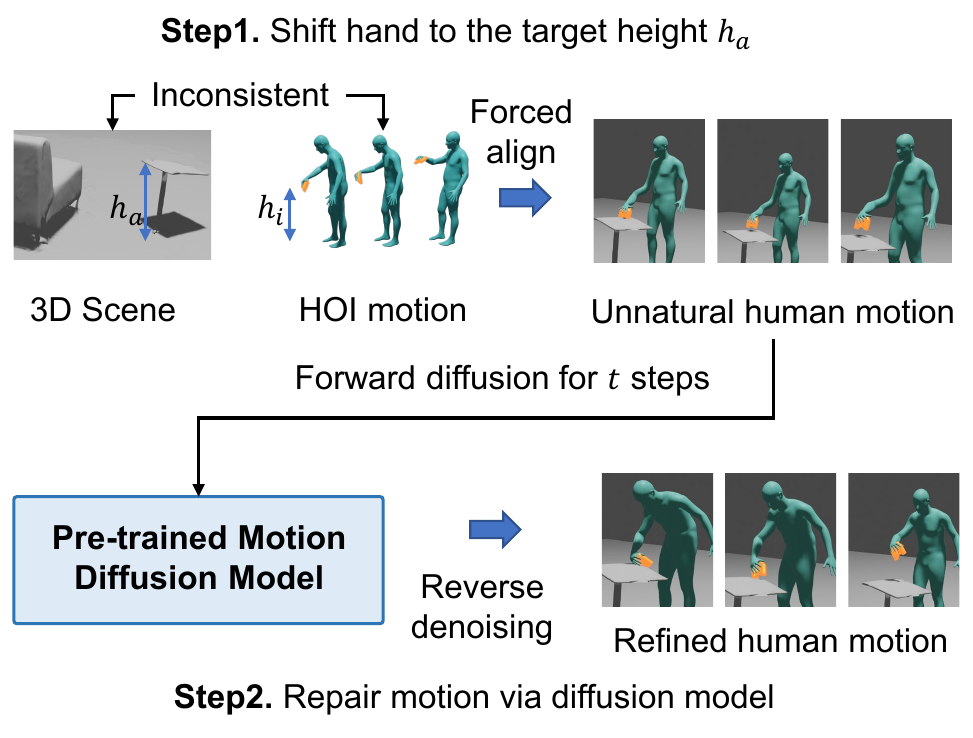}
\caption{\label{fig_dataset} Method of our motion alignment.}
\end{figure}

\begin{figure*}
\centering
\includegraphics[width=\textwidth]{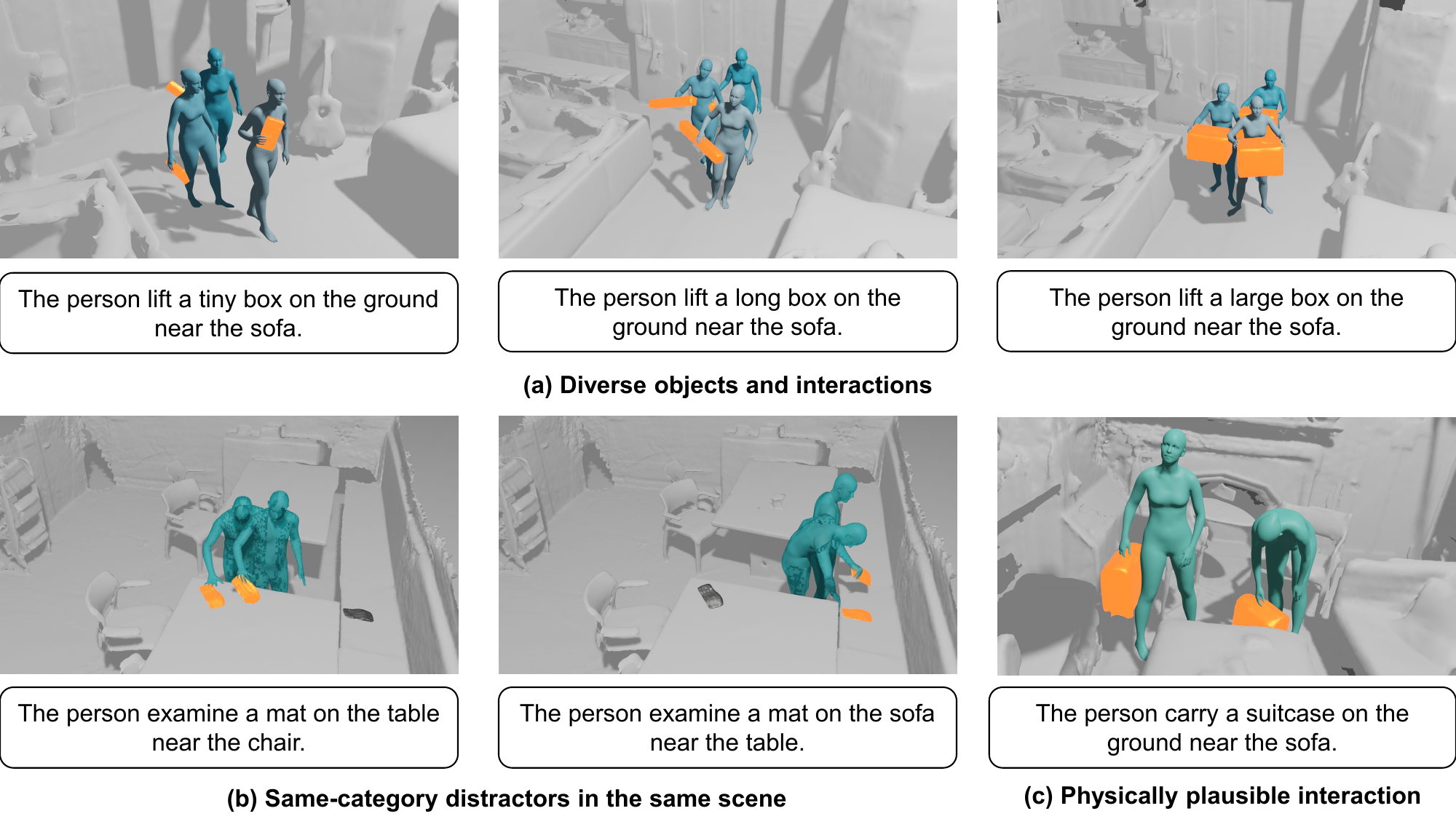}
\caption{\label{visdataset} Visualizations of our dataset.}
\end{figure*}

\begin{table*}[!htb]
    \centering
    \scalebox{1.0}{
    \begin{tabular}{ccccccc}
    \toprule
        Dataset & Scenes & Movable Objects & \makecell{Movable Object \\Categories} & \makecell{Annotated\\ Interaction Types} & Frames & \makecell{Nature Language \\Annotations} \\ 
        \midrule
        PiGraph\cite{pigraph} & 30  & $\times$ &-&-& 0.1M & $\times$ \\
        PROX-Q\cite{proxq} & 12  & $\times$ &-&-& 0.1M & $\times$ \\
        GTA-IM\cite{gtaim} & 49  & $\times$ &-& - & 1.0M & $\times$ \\
        CIRCLE\cite{mulframe1} & 9  & $\times$ &-&-& 4.3M & $\times$ \\
        HUMANISE\cite{humanise} & 643 & $\times$ &-&4& 1.2M & \checkmark \\
        
        TRUMANS\cite{trumans} & 100  & \checkmark & 20 & 10 & 1.6M & $\times$ \\
        \midrule
        Ours & 618 &  \checkmark & 71 & 21 & 2.2M & \checkmark \\
        \bottomrule
    \end{tabular}}
    
    \caption{\textbf{Comparisons of our dataset with other Human-Scene Interaction datasets.} }
    \label{tab_dataset}
\end{table*}

To enable text-controlled human-object interaction generation in 3D scenes with movable objects, we construct InteractMove, a novel dataset that enriches existing human-object interaction (HOI) data with realistic, richly annotated 3D scene contexts. Instead of collecting new data from scratch, which is costly and time-consuming, we automatically align existing motion sequences from BEHAVE~\cite{behave} and GRAB~\cite{grab} datasets with 3D scenes to achieve a scalable yet high-quality solution. Our construction process emphasizes the following key principles: (1) \textbf{Movable target objects}: Diverse objects are placed in semantically appropriate areas of the scene, including multiple distractors of the same category, to facilitate spatial understanding. (2) \textbf{Physically Coherent Motion Alignment}: Human motion sequences are adjusted to achieve realistic interactions with objects at different positions. (3) \textbf{Scene-aware Filtering for Physical Plausibility}: The aligned motion-scene pairs are filtered to remove cases violating physical constraints, such as foot-ground detachment, boundary overflow, or human-object collisions.

\subsection{Object Placement in 3D Scenes}
\label{objset}

We first collect object and human-object interaction data from existing HOI datasets BEHAVE~\cite{behave} and GRAB~\cite{grab}, totaling 71 object categories and 21 interaction types. 
Taking the interaction of \textit{take picture with camera} as an example, we will discuss the placement process here. To integrate them into realistic 3D environments, we utilize the ScanNet~\cite{scannet} dataset to obtain 3D scene and utilize the Sr3D dataset~\cite{sr3d} to obtain object-region annotations and relative spatial relations annotations in ScanNet. For each target interaction, we identify appropriate placement surfaces in the 3D scene. For instance, a camera might be located on the surface of a table. Sr3D\cite{sr3d} provides annotations for such regions and their relative spatial positions like \textit{a table next to a door}. We sample these surfaces within the scene where objects can be placed and ensure that their relative positions are annotated by Sr3D\cite{sr3d}. Then, based on the interaction label provided by the HOI dataset type and location annotations provided by Sr3D\cite{sr3d}, we can automatically generate the full interaction textual annotations from templates: \textit{A person takes pictures with the camera on a table next to a door}. For each scene, we also put multiple instances of the same category as the target object on every reasonable surface. 
For example, if there are $k$ placeable surfaces in the scene, we will randomly select a subset and put cameras on these surfaces when aligning an interaction of \textit{A person takes pictures with the camera}. This requires the model to learn language-conditioned object disambiguation, such as identifying \textit{The camera on a table next to a door}.

\subsection{Motion Alignment}
\label{motionalign}

One of the core challenges in aligning HOI data with new scene placements is the mismatch between the original object height and its new scene-constrained height (e.g., an object being on a shelf or table). As shown in Fig.~\ref{fig_dataset}, to ensure the interaction remains realistic, we adjust the corresponding human motion, particularly hand and arm movements, to match the new object location. We apply a motion inpainting strategy based on a pre-trained motion diffusion model~\cite{MDM}, focusing on editing the relevant hand joint trajectories during the phase when the hand comes into contact with the object.
Specifically, suppose the initial height of the object is $h_i$, and the adjusted height is $h_a$ (the height of the surface we plan to put the object on). We first identify the moment when the human hand initiates interaction with the object, characterized by two features: the object begins to exhibit motion, and the absolute position of the human hand is proximal to the object. Then, to align the motion with the scene, we first \textit{shift the human hand joints to the target height} $h_a$ as shown in Fig.~\ref{fig_dataset}: for the hand joint positions within the $T$ frames before and after the interaction start frame, we forcibly adjust the height based on the formula: $M_t = (h_a - h_i) \times \frac{t}{T}$. This ensures a smooth transition, with the height aligning with the new height when contacting the object. However, this forced adjustment can make the motion sequence unnatural. Then, we \textit{repair the motion using a pretrained diffusion model} as shown in Fig.~\ref{fig_dataset}: we apply forward diffusion for $t$ steps to introduce noise, and then leverage a pretrained motion diffusion model~\cite{MDM} to perform reverse denoising, progressively restoring realistic and coherent human motions.
This design maximizes the preservation of the majority of the interaction process, with only the approaching phase being modified, thus retaining the valuable original interaction data. 

\subsection{Physics-based Filtering and Validation}
\label{filtering}
After alignment, we apply strict filtering to ensure all motion-scene combinations are physically plausible and scene-aware. Our filtering criteria include: (1) \textit{Foot-ground contact}: Ensures the human maintains realistic contact with the ground; sequences with abnormal foot elevation or penetration are discarded. (2) \textit{Scene boundary constraints}: Motion sequences that move the human outside the visible scene space are rejected by monitoring the root joint trajectory. (3) \textit{Collision detection}: We compute distances between the human mesh and nearby objects or walls, removing samples with significant interpenetration. This filtering pipeline guarantees that each retained sequence is compatible with the geometry and physics of the scene, making the dataset suitable for training models on physically realistic 3D human-object interactions.

\subsection{Quantitative Statistics and Visualizations}
\label{statitics}

Our InteractMove dataset contains 30.5k interaction sequences across 618 richly annotated indoor 3D scenes, encompassing 71 different types of movable objects, such as cameras, apples, and mugs. Compared with existing Human-Scene Interaction (HSI) datasets, InteractMove exhibits three unique advantages, as visually illustrated in Fig.~\ref{visdataset}: (1) \textit{Diverse object types and interaction complexity}: As shown in Fig.~\ref{visdataset} (a), InteractMove supports a wide range of object categories with diverse object size and interaction strategies (e.g. using one hand or two hands). This significantly enriches the interaction patterns and poses new challenges for motion generation models to adapt to object size, shape, and affordances. (2) 
\textit{Multiple movable objects per scene}: As shown in Fig.~\ref{visdataset} (b), our scenes contain multiple interactable objects of the same category, placed in semantically reasonable locations. This setup introduces same-category distractors, requiring models to perform fine-grained spatial reasoning and accurate object grounding based on text. (3) \textit{Physically plausible interaction}: As shown in Fig.~\ref{visdataset} (c), our dataset includes data of humans manipulating objects with large-range movement. Thanks to our scene-aware motion adjustment and filtering pipeline, all motions in our dataset are collision-free and physically reasonable. 

As shown in Tab.~\ref{tab_dataset}, our dataset outperforms existing human-scene interaction datasets in terms of the number of scenes, scale of interaction frames, and variety of movable objects. Unlike prior datasets that often involve static furniture (e.g., beds or sofas), our InteractMove enables text-guided human-object interactions in 3D scenes with movable objects, along with semantic natural language descriptions. Also, the statistics in Tab.~\ref{tab_quality} also prove the quality of our synthesized data.

%% file: sec/4_method.tex
\section{Method}
\label{method}

\begin{figure*}[htbp]
\centering
\includegraphics[width=0.81\textwidth]{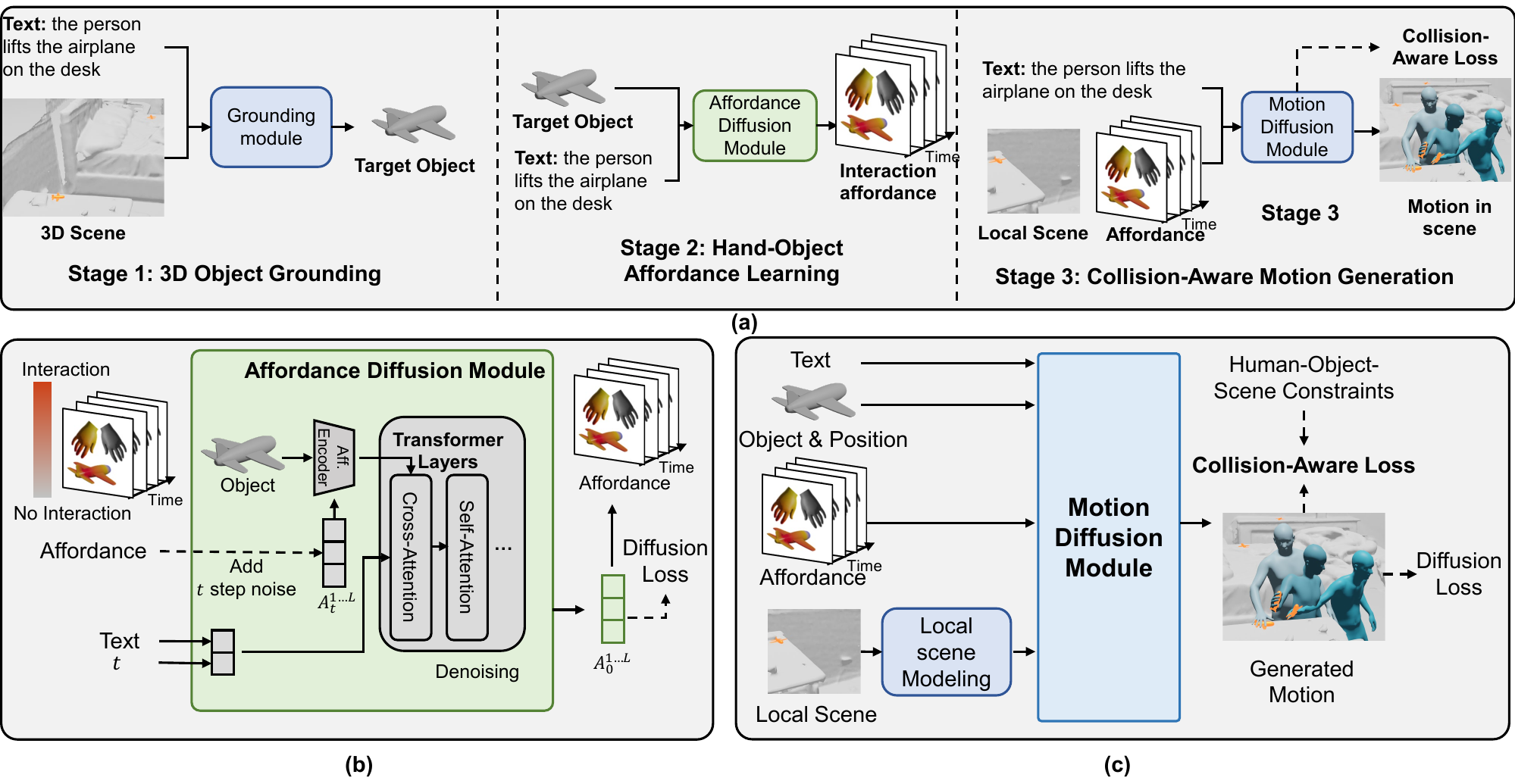}
\caption{\label{fig_method_overview}\textbf{Overview of the proposed framework.} 
(a) Given a text instruction, we first locate the interactive object via a pre-trained grounding model. Then, conditioned on the object point cloud and textual instruction, we generate hand-object affordances. Finally, a collision-aware motion generation module synthesizes human motion and object trajectory, incorporating local scene geometry and learned affordances.
(b) Hand-object affordance diffusion module. 
(c) Collision-Aware motion diffusion module.}
\end{figure*}

\begin{figure}[htbp]
\centering
\includegraphics[width=0.8\linewidth]{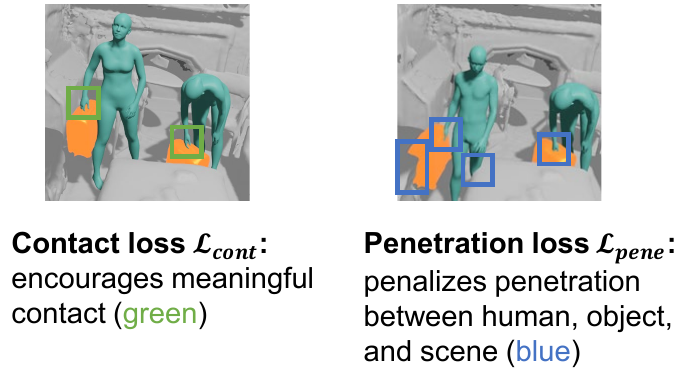}
\caption{\label{fig_method_loss} Our Collision-Aware Loss.}
\end{figure}

\subsection{Overview}

Our method addresses the task of text-conditioned human-object interaction generation in 3D scenes with moveable objects, producing human motion sequences $X$ and object trajectories $Y$ based on text $T$ and 3D scene information $S$ including a set of object point clouds $O\in \mathbb{R}^{N\times 3}$ for $N$ points of $M$ objects in the scene.

Compared to conventional Human-Scene or Human-Object Interaction generation tasks, our setting introduces unique challenges: the model must (1) identify the target object from free-form language in a 3D scene, (2) adapt the interaction to diverse object geometries and task descriptions, and (3) ensure the generated object trajectory is physically plausible and avoids collisions with the surrounding scene.
To tackle these challenges, we propose a novel \MethodName{} (\MethodNameShort{}) framework as shown in Fig.~\ref{fig_method_overview} (a). We begin with \textit{3D object grounding} using a pretrained grounding module~\cite{ZSVG3D} with the text condition $T$ to identify its point cloud $O'$ for the next stage.
Next, we perform \textit{hand-object affordance learning} uses an affordance diffusion module (more details are provided in Sec.~\ref{subsec_aff}), which takes the object point cloud and text instruction as inputs and generates a fine-grained hand-object joint affordance $A \in \mathbb{R}^{N\times J\times L}$, for $N$ points of the object, $J$ points of the hand, and $L$ frames, to guide plausible hand-object contact by considering the object shape and size. This affordance represents the likelihood of interactions occurring between hand joints and object surfaces over time and is used to guide the interaction motion generation, enabling more accurate interactions aligned with object size and interaction semantics. 
Finally, we incorporate a \textit{collision-aware motion generation} that voxelizes the region around the interactive object to evaluate spatial accessibility, as the local scene information around the object to be interacted with is more critical for preventing the object from collision through the scene. We also combined with a collision-aware loss that enforces physically plausible motion and prevents interpenetration. Conditioned on the text, local scene, and learned affordance, our model generates physically plausible motion sequences that align with both interaction semantics and environmental constraints.

\subsection{Preliminary: Diffusion Models}
\label{Diffusion}
We utilize the Denoising Diffusion Probabilistic Model (DDPM)~\cite{diffusion} to generate both hand-object affordance and motion sequences under conditioning. Given a ground truth signal $A_0$, the forward diffusion process adds Gaussian noise step-by-step:
\begin{equation}
q(A_t | A_{t-1}) = \mathcal{N}(A_t; \sqrt{1-\beta_t} A_{t-1}, \beta_t I),
\end{equation}
where $\beta_t$ is a noise schedule. The closed-form expression for $A_t$ is:
\begin{equation}
A_t = \sqrt{\bar{\alpha}_t} A_0 + \sqrt{1 - \bar{\alpha}_t} \epsilon, \quad \epsilon \sim \mathcal{N}(0, I),
\end{equation}
with $\bar{\alpha}_t = \prod_{s=1}^{t}(1 - \beta_s)$.
The reverse process is parameterized by a neural network $G_\theta$ which estimates $\hat{A}_0$ conditioned on input $A_t$ and context $c$:
\begin{equation}
\hat{A}_0 = G_\theta(A_t, c).
\end{equation}
The model is trained to minimize the mean squared error:
\begin{equation}
\mathcal{L}_{diff} = \mathbb{E}_{A_0,t} \left\| A_0 - \hat{A}_0 \right\|_2^2.
\label{diff_loss}
\end{equation}

\subsection{3D Object Grounding}
\label{subsec_grounding}
To determine which object the human should interact with, we first use a 3D visual grounding model (e.g., ZSVG3D~\cite{ZSVG3D}) to locate the object referenced in the input text $T$. The output is a selected target object point cloud $O' \in \mathbb{R}^{N \times 3}$.
Although we use ZSVG3D~\cite{ZSVG3D}, our pipeline is agnostic to the specific grounding module and can flexibly incorporate future grounding advancements.

\subsection{Hand-Object Affordance Learning}
\label{subsec_aff}
Objects in 3D environments vary greatly in their shapes, sizes, and potential interaction strategies. For example, interacting with a small cup may only require one hand, while lifting a heavy box might necessitate both hands, and the hand joints and object parts involved in object interaction also differ. To handle this diversity, our second stage, as shown in Fig.~\ref{method}(b), takes the object point cloud as inputs and models the fine-grained spatial relationship between human hands and object surfaces, providing interaction-aware guidance based on the hand-object joint affordances.

Given the object point cloud and text conditions, the model generates the hand-object joint interaction affordance. We calculate the distance between each point of the object and each joint of the human at each frame to get the distance map $d\in \mathbb{R}^{N\times J\times F}$, where $N$ is the number of points in the object, $J$ is the number of human joints, and $F$ is the number of frames. Then we normalize it as $C_{ijn}=\exp \left( -\frac{1}{2}\frac{d_{ijn}}{{\sigma}^2} \right)$ to assign a higher value to closer point-joint pairs, indicating their high relations. To differentiate interaction types involving single-hand control or bi-hand control, we compute individual affordance scores for each hand and establish a threshold $\tau$ to determine hand engagement status. Subsequent normalization based on $\tau$ ensures temporal continuity in the resultant affordance signals: $A_{ijn}=\mathbf{1}_{ C_{ijn}>\tau}\cdot \frac{C_{ijn}-\tau}{1-\tau}$. $A_{ijn}$ then indicates whether the $j-$th hand joint is involved in the interaction with the $i-$th point of the object in the $n-$th frame or not (if $A_{ijn}=0$).

To denoise the affordance using a diffusion model, we extract object features from the point cloud using PointNet\cite{pointtransformer} and fuse them with the noisy affordance via cross-attention. The fused features, along with timestep and text embeddings, form the input to a Transformer decoder, which predicts the final hand-object joint interaction affordance for the subsequent interaction generation stage. Same as Eq(\ref{diff_loss}), we use the diffusion loss to supervise the model training.

\subsection{Collision-Aware Motion Generation}
\label{subsec_interact}

Generating physically plausible interactions requires respecting the constraints imposed by the surrounding 3D scene. Thus, we design a collision-aware motion synthesis module guided by local scene modeling and a collision-aware loss.
As the local scene information around the object to be interacted with is more critical for preventing the object from collision through the scene, we propose a local scene understanding model that voxelizes the region around the interactive object to evaluate spatial accessibility, providing local scene information for the model. We also combined with a collision-aware loss that enforces physically plausible motion and prevents interpenetration. 

\textbf{Local Scene Modeling.} 
We voxelize the 3D scene into occupancy grids $\mathcal{S}' \in \mathbb{N}^{N_x \times N_y \times N_z}$, indicating whether each voxel is occupied. Around the target object, we extract a region and divide it into patches on the x-y plane. The feature for each patch is derived by pooling occupancy values along the z-axis. These 2D patch features are then encoded using a Vision Transformer (ViT) to obtain local scene feature tokens $f_{local}$ that inform motion synthesis.

\textbf{Collision-aware Loss.}
The unique challenge of our proposed task is the complexity of the entities involved, especially the dynamics of objects should be consistent with the scene and human motions. Therefore, we introduce a collision-aware loss function composed of two components as shown in Fig.~\ref{fig_method_loss}: contact loss $\mathcal{L}_{\text{cont}}$ and penetration loss $\mathcal{L}_{\text{pene}}$. The contact loss encourages meaningful contact between hand joints and object surfaces, while the penetration loss penalizes any interpenetration between human body parts, the object, and the scene geometry. Together, these terms enforce physical plausibility and consistency in the generated interactions.
The contact loss $\mathcal{L}_{cont}$ is formed as:
  \begin{equation}
\mathcal{L}_{cont}=||d(\tilde{j}_{\mathrm{}},\hat{p}_{\mathrm{obj}}^{\mathrm{}})||^2,
  \label{eq:loss1}
\end{equation}
where $\tilde{j}$ indicates human joints within a distance threshold from the target object and $\hat{p}_{\mathrm{obj}}^{\mathrm{}}$ indicates the object points closest to these human joints. The penetration loss $\mathcal{L}_{pene}$ is formed as:
  \begin{equation}
\mathcal{L}_{pene}=||d(\tilde{v}_{\mathrm{}},\hat{p}_{\mathrm{obj}}^{\mathrm{'}})||^2,
  \label{eq:loss2}
\end{equation}
where $\tilde{v}$ indicates human vertices that penetrate the object surface, and $\hat{p}_{\mathrm{obj}}^{\mathrm{'}}$ indicates the object points closest to these human vertices.

Considering the necessity of keeping object movement aligned with the scene, we introduce a test-time-penetration constraint. During the denoise process, we recover the human and object point cloud based on the denoise results on step $t$, and move it along the negative-gradient direction of the test-time penetration loss $\mathcal{L}_{ttp}$. We first filter the vertex set where penetration occurs:
\begin{equation}
\mathcal{P}=\{(i,j) | -\mathbf{n}_j^\mathrm{T}\cdot(\mathbf{V}_{gen}^i-\mathbf{V}_{scene}^j)>0\},
  \label{eq:pene}
\end{equation}
where $\mathbf{V}_{gen}^i$ is the vertex set of the recovered mesh, $\mathbf{V}_{gen}^i$ is the nearest scene vertex set, and $\mathbf{n}_j^\mathrm{T}$ is the local normal vector. Then we calculate the test-time penetration loss:
\begin{equation}
\mathcal{L}_{ttp}=\sum_{(i,j)\in\mathcal{P}}{||\mathbf{V}_{}^i-\mathbf{V}_{}^j||_2}.
  \label{eq:pene2}
\end{equation}

\textbf{Interaction Denoising.} 
We use a diffusion module to generate the interactions. 
We apply positional encoding to the noisy interaction to obtain interaction features $F_{int}$. 
The interaction affordance and the object features are concatenated and fed into the fully connected layers to obtain the $F_{obj}$. All condition tokens, including $F_{int}$, $F_{obj}$, and the local scene feature $F_{local}$, are concatenated and passed through a transformer encoder to produce the denoised output $\hat{Y}$. The model is trained with a total loss:
\begin{equation}
    \mathcal{L}_{total} = \mathcal{L}_{diff} + \lambda_1 \mathcal{L}_{cont} + \lambda_2 \mathcal{L}_{pene}
\end{equation}
where $\mathcal{L}_{diff}$ is the diffusion reconstruction loss same as Eq(\ref{diff_loss}), $\mathcal{L}_{cont}$ and $\mathcal{L}_{pene}$ are collision-aware losses, and $\lambda_1,\lambda_2$ are hyper-parameters. $\mathcal{L}_{ttp}$ is only applied for inference, inhibiting the penetration after every step of the denoising.

%% file: sec/5_exp.tex
\section{Experiments}
\label{sec_exp}

\begin{table}[t]
    \centering
    \scalebox{0.8}{
    \begin{tabular}{cccccccccc}
    \toprule
        Method & \makecell{Goal \\Distance$\downarrow$} & \makecell{Multi-\\modality$\uparrow$} & \makecell{Physical\\Realism$\uparrow$} & \makecell{Non-collision \\Score$\uparrow$}   \\ \hline
        MDM\cite{MDM} & $0.904$ &  $1.33$ & $ 0.474$ & $84.97$   \\ 
        HUMANISE\cite{humanise} & $0.847$ &  $1.17$ & $ 0.659$ & $95.21$   \\ 
        GOAL\cite{goal} & $0.820$ & $1.25$ & $0.708$ & $96.63$   \\  \midrule
        Ours & $\bm{0.791}$ &  $\bm{1.58}$ & $\bm{0.813}$ & $\bm{98.36}$   \\ 
        \bottomrule
    \end{tabular}}
    \caption{Quantitative evaluations on our dataset. 
    }\label{tab_main}
\end{table}

\begin{table}[t]
    \centering
    \scalebox{1.0}{
    \begin{tabular}{cccccccccc}
    \toprule
        Method & \makecell{Multi-\\modality$\uparrow$} & \makecell{Physical \\Realism$\uparrow$} &\makecell{Non-collision \\Score$\uparrow$}   \\ \hline 
        TRUMANS\cite{trumans}  & $1.29$ & $0.707$ & $98.73$   \\  \midrule
        Ours & $1.33$ & $0.754$ & $99.03$   \\ 
        \bottomrule
    \end{tabular}}
    \caption{Quantitative evaluations on the TRUMANS dataset. For fairness, we conduct the comparison only on samples involving interactions with movable objects.
    }\label{tab_tumans}
\end{table}

\begin{table}[t]
    \centering
    \tabcolsep=1mm
    \scalebox{0.8}{
    \begin{tabular}{ccc|cccc}
    \toprule
        \makecell[c]{Grounding \\Module} & \makecell[c]{Hand-Object\\ Affordance} & \makecell[c]{Local-Scene \\Modeling} & \makecell[c] {Goal \\Dist.$\downarrow$} & \makecell[c] {Multi-\\modality$\uparrow$} & \makecell[c] {Physical \\Realism$\uparrow$} & \makecell[c] {Non-\\collision$\uparrow$}   \\ \hline

        $\times$ & $\times$ &  \checkmark & $1.545$ & $1.73$ & $0.464$ & $90.18$   \\

        \checkmark & $\times$ & $\times$ & $0.895$ & $1.69$ & $0.524$ & $78.36$   \\
        
        \checkmark & $\times$ &  \checkmark & $0.793$ & $1.47$ & $0.570$ & $95.21$   \\

        \checkmark & \checkmark & $\times$ & $0.803$ & $\bm{1.87}$ & $0.752$ & $80.24$   \\
        
        \checkmark & \checkmark &  \checkmark & $\bm{0.791}$ &  ${1.58}$ & $\bm{0.813}$ & $\bm{98.36}$   \\ 
        
        \bottomrule
    \end{tabular}}
    \caption{Ablations of each component in our method.
    }
    \label{tab_abla}
\end{table}

\begin{table}[t]
    \centering
    \tabcolsep=2mm
    \scalebox{0.85}{
    \begin{tabular}{ccc|cccc}
    \toprule
        \makecell[c]{$\mathcal{L}_{cont}$} & \makecell[c]{$\mathcal{L}_{pene}$} & \makecell[c]{$\mathcal{L}_{ttp}$} &\makecell[c] {Goal \\Dist.$\downarrow$} & \makecell[c] {Multi-\\modality$\uparrow$} & \makecell[c] {Physical \\Realism$\uparrow$} & \makecell[c] {Non-\\collision$\uparrow$}  \\ \hline
        
        $\times$ & $\times$ &  $\times$ & $0.803$ & $\bm{1.87}$ & $0.752$ & $80.24$   \\
        
        $\times$ & $\times$ &  \checkmark & $0.793$ & $1.55$ & $0.754$ & $96.15$   \\

        $\times$ & \checkmark & $\times$ & $0.796$ & $1.72$ & $0.775$ & $90.02$   \\
        
        \checkmark & $\times$ & $\times$ & $0.798$ & $1.76$ & $0.788$ & $87.15$   \\

        \checkmark & \checkmark & $\times$ & $0.796$ & $1.66$ & $0.808$ & $95.44$   \\
        
        \checkmark & \checkmark &  \checkmark & $\bm{0.791}$ &  ${1.58}$ & $\bm{0.813}$ & $\bm{98.36}$   \\ 
        
        \bottomrule
    \end{tabular}}
    \caption{Ablations of the collision-aware loss.     }
    \label{tab_loss}
\end{table}

\begin{table}[t]
    \centering

    \scalebox{1.0}{
    \begin{tabular}{ccccc}
    \toprule
        \makecell[c]{Number of\\Instances} & \makecell{Goal \\Dist.$\downarrow$} & \makecell{Multi-\\modality$\uparrow$} & \makecell{Physical \\Realism$\uparrow$} & \makecell{Non-collision \\Score$\uparrow$}   \\ \hline

        Unique & $0.405$ & $1.47$ & $ 0.819$ & $98.77$   \\
        
        Multiple  & $0.793$ & $1.60$ & $ 0.805$ & $98.21$   \\  
        
        \bottomrule
    \end{tabular}}
    \caption{Experiments on the number of instances of the same category as the target object.
    }
    \label{tab_num}
\end{table}

\begin{table}[t]
    \centering
    \scalebox{0.9}{
    \begin{tabular}{ccccc}
    \toprule
        Motions & FID$\downarrow$ & Diversity$\uparrow$ & \makecell[c]{Physical\\ Realism$\uparrow$}   \\ \hline

        Original Motion  & $0$ & $0.8629$ & $ 0.8327$ \\ 
        Forced Alignment &$0.670$& $0.8531$& $0.8186$\\
        Refined Motion  & $0.133$ & $0.8459$ & $ 0.8213$ \\ 
        
        \bottomrule
    \end{tabular}}
    \caption{Ablation of our dataset construction. We evaluate the quality of the original HOI motion and our aligned motion (aligned with the 3D scene).}
    \label{tab_quality}
\end{table}

\begin{figure}
  \centering
  \includegraphics[width=0.75\linewidth]{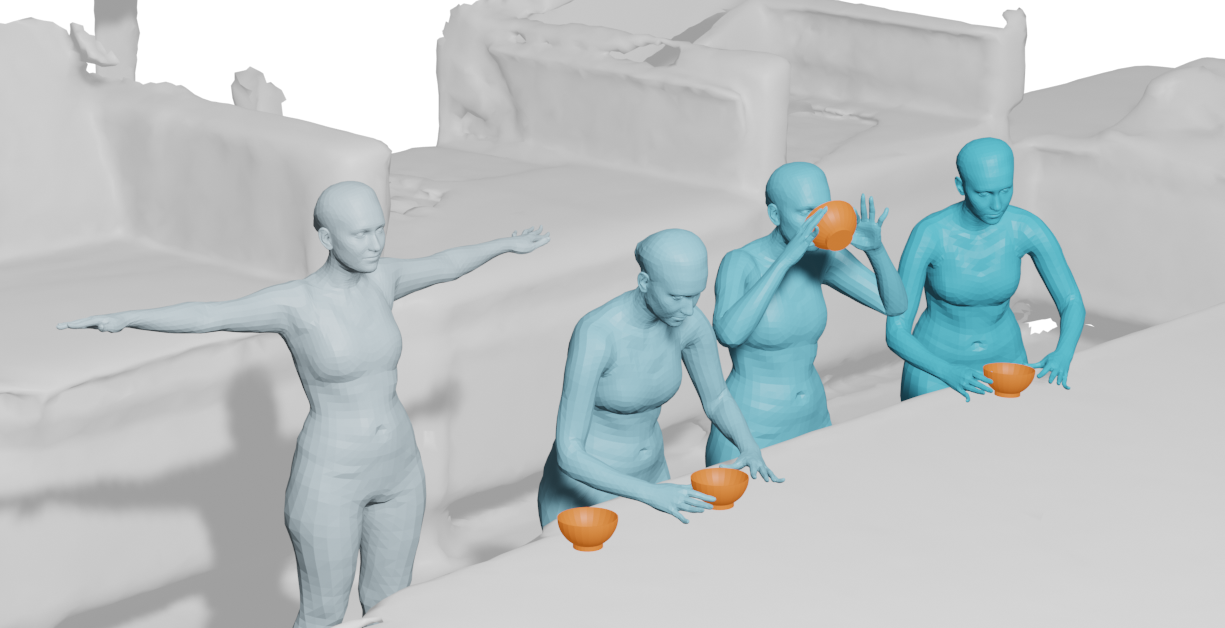}
  \caption{Visualization. The prompt is \textit{The person drinks the bowl on the desk near the sofa.}}
  \label{fig:addvis}
\end{figure}
\subsection{Evaluation Metrics}
We adopt the following metrics to evaluate interaction quality:
\textit{1) Goal Distance (Goal Dist).} Measures how well the human interacts with the target object, computed as the average minimum distance between the human body and object surfaces over time. Lower is better.
\textit{2) Multimodality.} Assesses the diversity of actions generated from the same prompt and scene. Defined as the average $\mathcal{L}_2$ distance between multiple generated motions. Higher is better.
\textit{3) Physical Realism.} Evaluates whether the motion appears physically plausible, using a pre-trained model to score each frame as realistic (1) or not (0). The final score is the average over all frames. Higher is better.
\textit{4) Non-collision Score.} Measures the proportion of frames without collisions or penetrations. Higher is better.

\subsection{Quantitative Results}
We present the quantitative results in our InteractMove dataset in Tab.\ref{tab_main}. We can observe that: 
\textit{1) Goal Distance:} Our method achieves the best Goal Distance performance, demonstrating its ability to generate accurate interactions with the correct target object.
\textit{2) Physical Realism:} The results show our advantages in the physical plausibility of interactions, which we attribute to the joint modeling of hand-object affordance.
\textit{3) Non-collision Score:} Our method yields fewer collisions with the scene, confirming the effectiveness of our collision-aware motion generation design.
\textit{4) Multimodality:} Our approach achieves significantly higher diversity compared to previous methods, while still satisfying other constraints, indicating strong capability in generating diverse yet plausible interactions.

We further evaluate our method on the TRUMANS dataset~\cite{trumans} in Tab.~\ref{tab_tumans}. Unlike our dataset, TRUMANS includes only 20 interactive objects and 10 predefined interaction types, with discrete action labels instead of free-form language descriptions. While our method is designed for text-controlled interaction motion generation, it is also compatible with label-based inputs. To fit our task, we conduct the comparison only on samples involving interactions with movable objects. Since the TRUMANS dataset provides the target object location, we omit the Goal Distance metric. As shown in Tab.~\ref{tab_tumans}, using our evaluation metrics, our method still achieves higher scores in Physical Realism, Non-collision, and Multi-modality, validating the effectiveness of our affordance-based motion generation even on limited-action datasets.

\subsection{Ablation Studies}
\label{subsec_abla}
We conduct ablation studies to evaluate both our method and dataset construction.

\textbf{Ablations on Pipeline Components.} 
Tab.\ref{tab_abla} shows results after disabling key modules in our pipeline: 1) Without the grounding module, the model struggles to locate target objects and interaction regions, leading to a sharp drop in Goal Distance. 2) Removing the hand-object joint affordance module significantly reduces interaction realism. This is because the hand-object joint affordance provides fine-grained spatiotemporal guidance for interactions and offers unique conditions for different types of objects. 3) Without local scene modeling, predicted motions often collide with the environment, showing that scene constraints are crucial for spatial consistency. These experiments demonstrate the effectiveness of the proposed modules.

\textbf{Ablations on Collision-aware Loss.} Tab.\ref{tab_loss} compares three collision-aware losses.
The inclusion of both $\mathcal{L}_{cont}$ and $\mathcal{L}_{pene}$ as training-phase supervisory terms moderately reduced the Goal Distance, indicating their effectiveness in optimizing spatial positioning during human-object interactions. Constraint $\mathcal{L}_{cont}$ demonstrated greater efficacy in enhancing Physical Realism, while constraint $\mathcal{L}_{pene}$ more substantially improved the Non-collision Score, confirming their respective functional priorities: interaction assurance and collision prevention. As an inference-phase constraint, $\mathcal{L}_{ttp}$ achieved the most significant reduction in intersection artifacts through remarkable Non-collision Score improvement. All constraints exhibited measurable reductions in Multimodality, which we consider an essential trade-off between stringent safety requirements and behavioral diversity preservation.

\textbf{Distractor Impact.} Our dataset contains multiple interactable objects of the same category, requiring models to perform fine-grained spatial reasoning and accurate object grounding based on text. We study the impact of the number of same-category distractors within the scene on the model's final performance. The results are in Tab.\ref{tab_num}. The task is much harder when multiple distractors exist in the scene, demonstrating that our proposed dataset and task are non-trivial.

\textbf{Dataset Construction.} 
We evaluate motion quality to assess the effectiveness of our motion alignment techniques, as shown in Tab.\ref{tab_quality}. Original Motion denotes unmodified HOI motions from GRAB and BEHAVE; Forced Alignment refers to forcing aligning these motions to the scene without refinement; Refined Motion is our proposed motion alignment method. Results show that Refined Motion significantly improves motion quality over Forced Alignment, demonstrating its ability to preserve naturalness and reduce artifacts, thus validating the rationale behind our dataset construction strategy.

\textbf{Visualizations.} We provide a visualization showing several frames of the interaction generated by our method in Fig.\ref{fig:addvis}. The person lifts the bowl, drinks and puts it back on the desk without collision and correctly uses both hands.

%% file: sec/6_conclusion.tex
\section{Conclusions}
\label{conclusion}

In this paper, we introduce a novel task of text-controlled human-object interaction generation in 3D scenes with movable objects, and build the InteractMove dataset to support it. Our proposed pipeline, integrating 3D visual grounding, joint affordance learning, and collision-aware motion generation, effectively handles object identification, diverse interaction prediction, and generation of physically realistic motion. Experiments show that our method outperforms existing approaches in generating physically plausible and text-compliant interactions.

\noindent\textbf{Acknowledgements.}This work was supported by the grants from the National Natural Science Foundation of China 62372014, Beijing Nova Program, Beijing Natural Science Foundation 4252040 and the State Key Laboratory of General Artificial Intelligence, BIGAI, Beijing, China.